\documentclass[a4paper, 10pt, conference]{ieeeconf}
\IEEEoverridecommandlockouts                             
\usepackage{color}
\usepackage{graphicx} 
\usepackage{mathtools} 
\usepackage{siunitx}
\title{\LARGE \bf Soft Air Pocket Force Sensors for Large Scale Flexible Robots}

\author{Michael R. Mitchell, Ciera McFarland, and Margaret M. Coad
\thanks{The authors are with the Department of Aerospace and Mechanical Engineering, University of Notre Dame, Notre Dame, IN 46556, USA. {\tt\small \{mmitch23,cmcfarl2,mcoad\}@nd.edu}}%
}

\begin{document}

\maketitle
\thispagestyle{empty}
\pagestyle{empty}

\begin{abstract}
Flexible robots have advantages over rigid robots in their ability to conform physically to their environment and to form a wide variety of shapes. Sensing the force applied by or to flexible robots is useful for both navigation and manipulation tasks, but it is challenging due to the need for the sensors to withstand the robots' shape change without encumbering their functionality. Also, for robots with long or large bodies, the number of sensors required to cover the entire surface area of the robot body can be prohibitive due to high cost and complexity. We present a novel soft air pocket force sensor that is highly flexible, lightweight, relatively inexpensive, and easily scalable to various sizes. Our sensor produces a change in internal pressure that is linear with the applied force. We present results of experimental testing of how uncontrollable factors (contact location and contact area) and controllable factors (initial internal pressure, thickness, size, and number of interior seals) affect the sensitivity. We demonstrate our sensor applied to a vine robot---a soft inflatable robot that ``grows" from the tip via eversion---and we show that the robot can successfully grow and steer towards an object with which it senses contact. 
\end{abstract}

\section{Introduction}
Soft and flexible robots have advantages over traditional rigid robots in applications where the ability to conform physically to the environment and undergo large shape changes is critical. For example, a flexible continuum robot can conform to the shape of the nasal cavity to reach the brain and resect a tumor~\cite{burgner2013telerobotic}, and a soft inflatable truss robot can change its shape from a deployable package to locomote and manipulate objects~\cite{usevitch2020untethered}. While rigid robots typically require touch sensing to achieve safe and effective physical interaction, the passive or ``embodied" softness of flexible robots often allows them to interact physically with their environment in a desirable way without requiring the use of touch sensors~\cite{hao2016universal}. Nevertheless, the ability to sense the magnitude and direction of contacts along a flexible robot's surface is useful for tasks such as active adjustment of the force applied by the robot during manipulation, updates to the robot's control based on the contact state during navigation, or measurement of environment properties~\cite{shih2020electronic}.

The integration of touch sensors into soft and flexible robots is challenging, due to the need for the sensors to withstand large shape changes and sense information from an infinite-degree-of-freedom contact state. Estimated contact forces can be determined by comparing the measured robot shape to a model of the expected robot shape given known actuation inputs~\cite{thuruthel2019soft}, but the accuracy of this approach depends on having good models and data processing algorithms. Various touch sensing skins that sense force directly have been developed for soft and flexible robots, using working principles such as resistance change of liquid metal traces upon contact~\cite{dickey2017stretchable}, but they tend to be challenging to fabricate and integrate over large areas. Air pressure sensing is a promising technique for touch sensing of soft and flexible robots, because it allows the creation of lightweight sensors that provide a straightforward relationship between contact force and sensed pressure and can be integrated over large areas. Soft air pressure sensors have been demonstrated for rigid robot fingertips~\cite{kim20153d, alspach2018design}, arm modules~\cite{ alspach2018design, kim2018soft}, and skins~\cite{gruebele2021stretchable}, as well as for single-degree-of-freedom sensor-actuator modules~\cite{buso2020soft, JonesIROS2022}, but they have not yet been adapted for touch sensing in a large scale flexible robot.

\begin{figure}[tb]
\centerline{\includegraphics[width = \columnwidth]{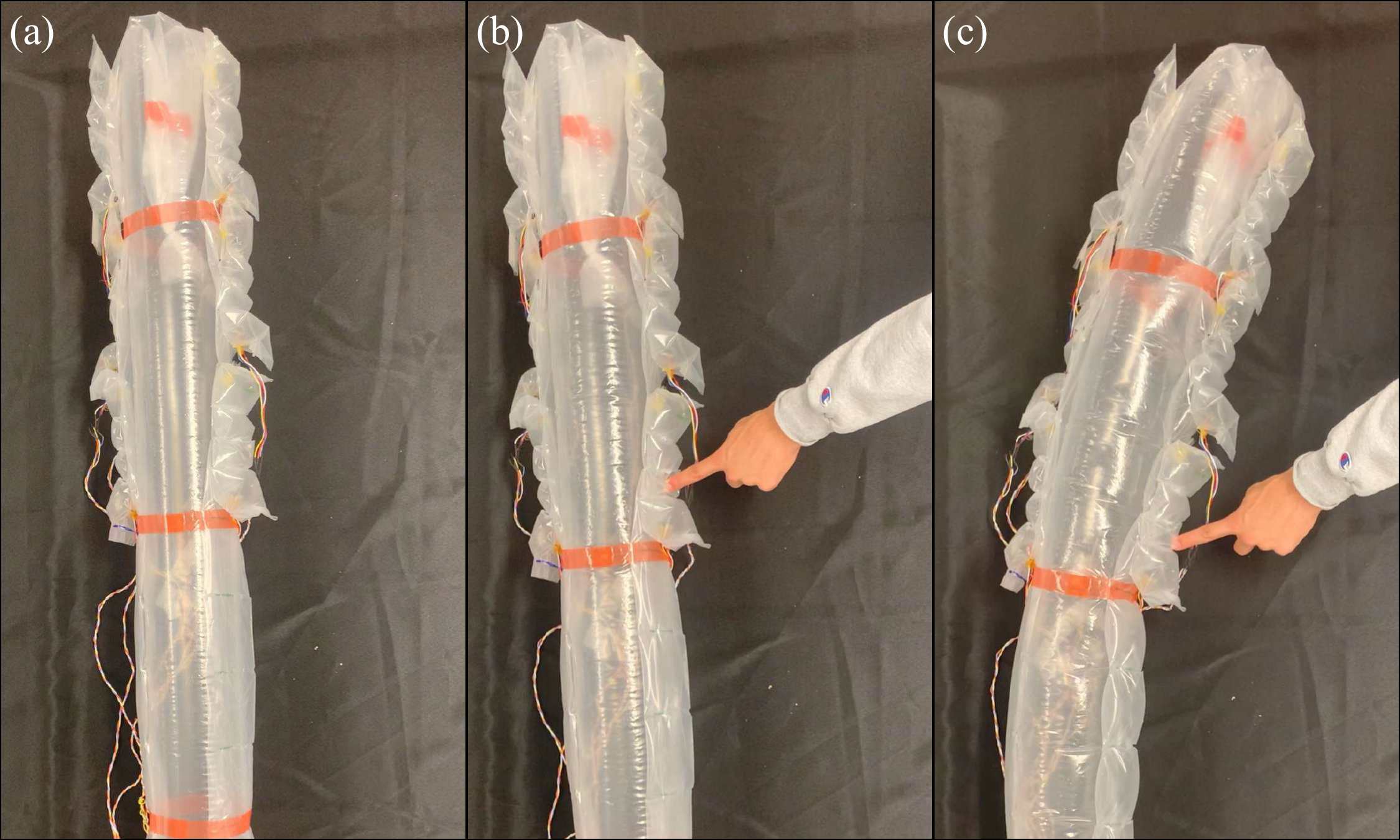}}
\caption{Demonstration of our soft air pocket force sensors used to control a vine robot to move towards human touch. (a) The robot is straight before force is applied. (b) Force is applied on one side of the robot. (c) The robot steers itself towards the source of the force.}
\vspace{-0.5cm}
\label{glamorshot}
\end{figure}

Vine robots~\cite{blumenschein_design_2020}, soft inflatable robots that ``grow" from the tip, are one type of soft flexible robot with a large surface area over which touch sensing is desirable. These robots are made of a tube of thin, flexible fluid-tight material that is folded inside of itself in its pre-grown state. To achieve growth, air pressure is typically used to inflate the robot and turn its body inside-out at the tip (called ``eversion"). These robots can extend to at least 72~m long~\cite{hawkes_soft_2017} and have been used to navigate confined spaces such as rocky tunnels in an archeological site~\cite{coad2019vine}. Adding touch sensing would allow these robots to navigate spaces more intelligently, manipulate objects with their whole body, and sense information about the stiffness properties of their environment. Thus far, distributed orientation, temperature, and humidity sensors have been integrated onto a vine robot using flexible sensor bands~\cite{GruebeleRoboSoft2021}, and tactile perception has been achieved using resistive curvature sensors to measure the robot's shape and calculate the applied force~\cite{BryantIROS2022}, but direct force sensing has not been implemented on a vine robot.

In this work, we explore the capabilities of air pressure-based sensing to enable touch sensing for large scale flexible robots such as vine robots (Fig.~\ref{glamorshot}). We present a novel soft force sensor made of a sealed air pocket containing an internal air pressure sensor. Our force sensor is fabricated by sealing a largely non-stretchable thermoplastic membrane at the edges using an impulse heat sealer, with partial seals throughout its body to increase its flexibility. Because it is made of a thin membrane filled with air, the sensor is lightweight and feasible for integration onto inflatable robots. The sensor's cost is determined primarily by the cost of the internal air pressure sensor, making it relatively inexpensive. By adjusting the dimensions of the membrane, the sensor can be scaled up or down as desired to cover a large scale robot with the desired resolution. In the sections that follow, we describe the design and fabrication of the sensor, the results of experiments showing the effect of various factors on the sensitivity, and a demonstration of the sensors sensing contact of a vine robot while it grows and steers.

\section{Design} \label{Design}

In this section, we discuss our design objectives and the design and fabrication of our sensors, as well as how we integrated them on a vine robot.

\subsection{Design Objectives} \label{Design Objective}
We considered several design objectives in creating our sensing system. First, the sensors should be able to be attached to the body of a flexible robot and detect forces applied along its entire surface. Second, the sensors should have as high accuracy as possible and a spatial resolution that can be chosen depending on the desired application (higher resolution improves localization of applied forces but requires more individual sensors). Third, the sensors should be able to withstand the forces the robot exerts and the shape change the robot undergoes without inhibiting their function.
Fourth, the sensors should not hinder the robot's ability to move and apply forces; for use with vine robots, this means that the sensing system should be able to evert with the vine robot body and undergo the curvatures that the robot achieves as it steers.


\subsection{Design Overview} \label{Sensing System Design}

\begin{figure}[bt]
\centerline{\includegraphics[width = \columnwidth]{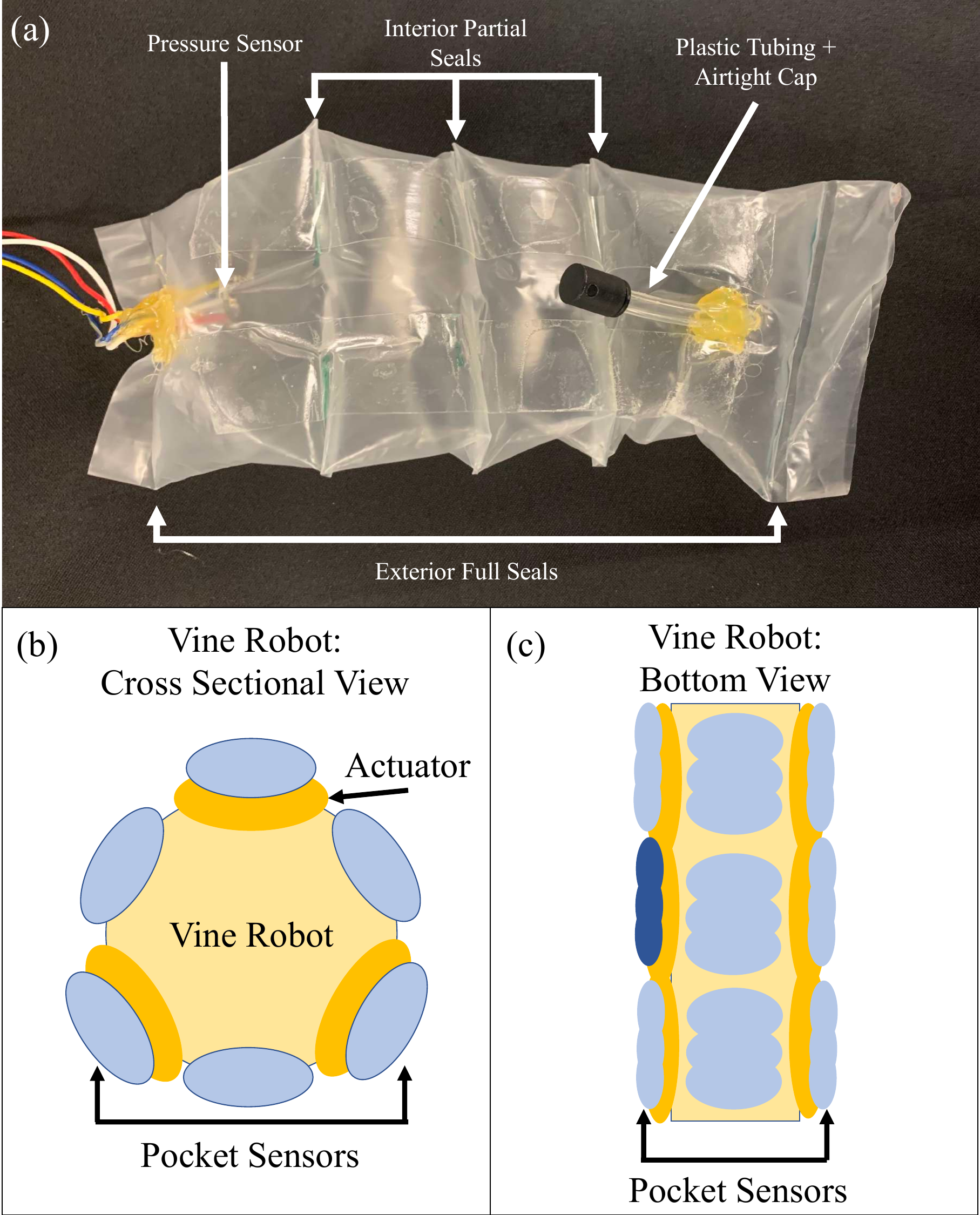}}
\caption{Design of our soft air pocket force sensor, and a possible sensor layout on a vine robot body. (a) The sensor is constructed from a large flexible airtight plastic tube that is fully sealed on both ends and inflated to make a pocket. Optional partial seals along the pocket decrease its thickness and increase its flexibility while still allowing internal air flow. An air pressure sensor is embedded inside the pocket. Optional small plastic tubing with an airtight cap allows adjustment of initial pocket pressure.
(b) For a vine robot with three actuators, one possible sensor layout has three sensors circumferentially on top of the actuators and three between the actuators.
(c) With two actuators running the length of the bottom, one possible sensor layout has nine sensors along the bottom, one of which is highlighted in dark blue.  
}
\vspace{-0.5cm}
\label{pocket}
\end{figure}

Our sensing system is shown in Fig.~\ref{pocket}. The force sensor (Fig.~\ref{pocket}(a)) is comprised of a pressurized, airtight pocket that has exterior full seals on its edges and interior partial seals throughout its body to increase its flexibility while still allowing internal airflow. An off-the-shelf air pressure sensor is embedded inside the pocket, such that forces applied anywhere on the surface of the pocket cause an increase in the pocket's internal air pressure, which can be read by the sensor. An optional tube with a removable cap can be attached to the sensor for adjusting the initial internal pressure; otherwise, the pocket can be permanently sealed after inflation to an initial internal pressure.

As one potential use case, an array of force sensors could be attached on the exterior of a vine robot's body, with sensors arranged radially around its circumference (Fig.~\ref{pocket}(b)) and distributed along its length (Fig.~\ref{pocket}(c)) with the desired resolution. One method of steering a vine robot involves the attachment of pneumatic artificial muscle actuators along the outside of the main growing tube~\cite{coad2019vine, greer2019soft, naclerio2020simple}; at least three actuators must be arranged around the robot's circumference in order to control the 3D position of the robot tip. For a vine robot steered in this way, to ensure that the force sensors will be the first point of contact with the environment, they could be attached on top of the actuators, and if additional circumferential resolution is desired, also between the actuators.


\subsection{Sensor Fabrication} \label{Pocket Manufacturing}

\begin{figure}[t]
\centerline{\includegraphics[width = \columnwidth]{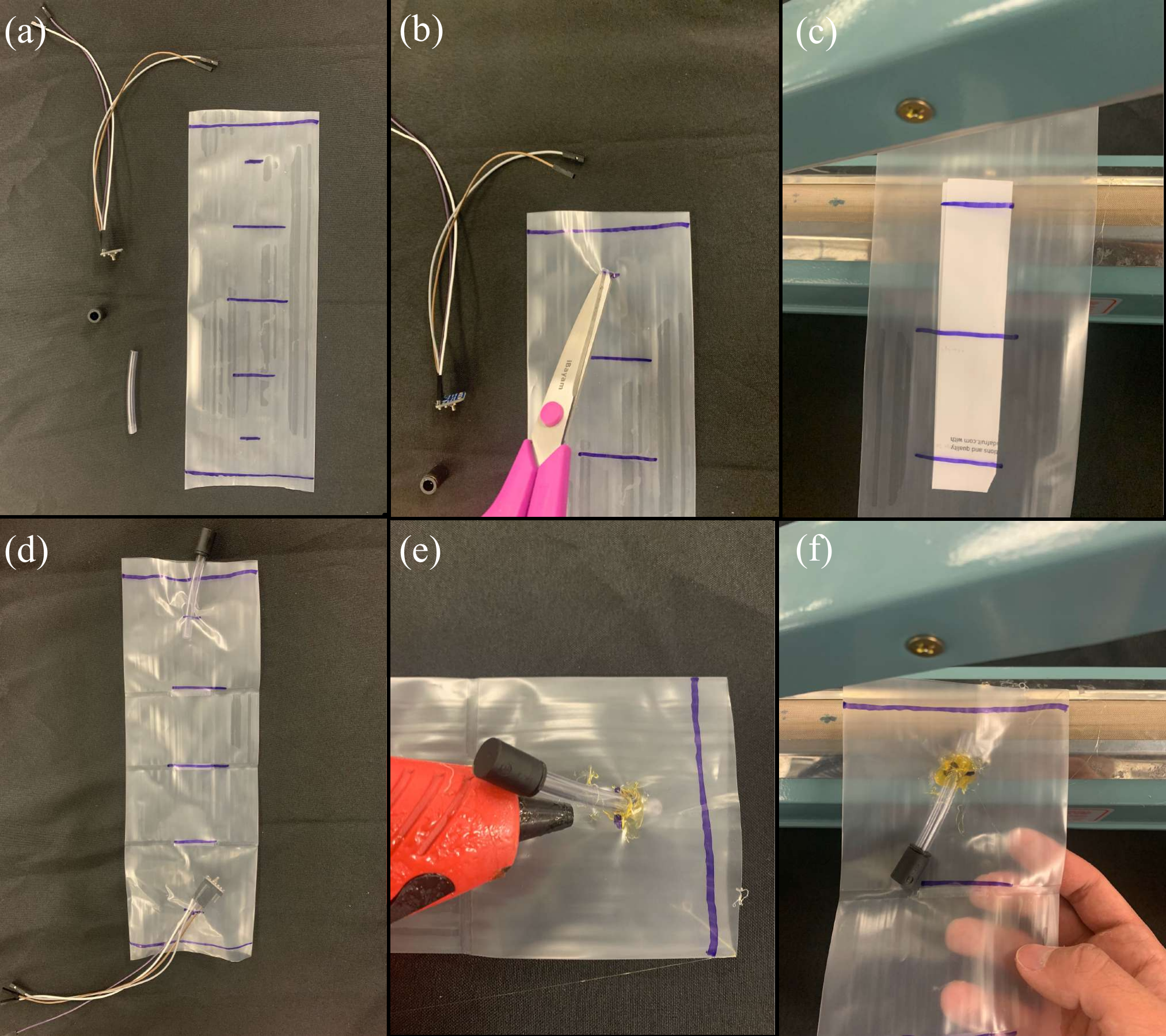}}
\caption{Fabrication steps for our force sensor. (a) The sensor components are the large plastic tubing with marks at the locations of incisions and seals, the air pressure sensor with wires, the small plastic tubing, and the airtight cap. (b) Make incisions at the smallest marks on the large plastic tubing. 
(c) Insert the paper into the middle of the large plastic tubing to prevent sealing in that area, and use an impulse heat sealer to create the interior partial seals.
(d) After removing the paper, insert the pressure sensor and the small plastic tubing with the airtight cap into the incisions. (e) Close the incisions using hot glue, both on the inside and outside of the large plastic tubing. (f) Make the exterior full seals using the impulse heat sealer. The pocket can then be inflated through the small plastic tubing.}
\vspace{-0.5cm}
\label{manufacturing}
\end{figure}

The fabrication of the force sensor is shown in Fig.~\ref{manufacturing}. The body of the sensor is made of a tube of low-density polyethylene (LDPE) plastic with 10.2~cm lay flat diameter and a different thickness and length depending on the sensor, as discussed later. The other sensor components include the air pressure sensor (MPRLS, Adafruit New York, NY) with attached wires, the air tube made of polyurethane plastic with 0.64~cm outer diameter, and an airtight push-to-connect tube cap with 0.64~cm inner diameter. 
To fabricate the sensor, we first marked the locations on the LDPE tubing where we planned to make the interior and exterior seals and the incisions for the sensor wires and air tube (Fig.~\ref{manufacturing}(a)). We then used scissors to cut the incisions for the sensor wires and the air tube (Fig.~\ref{manufacturing}(b)). Next, we used an impulse heat sealer with a heating element width of 5~mm to make the interior partial seals (Fig.~\ref{manufacturing}(c)). 
We placed paper with width approximately one third of the pocket's lay flat diameter inside the LDPE tube, such that the impulse heat sealer would seal together only the plastic not covered by the paper; after sealing, we removed the paper. We then placed the sensor and air tube inside the pocket (Fig.~\ref{manufacturing}(d)). We used a hot glue gun on its low temperature setting to glue the wires and air tube to the LDPE tube from both the interior and exterior of the incisions (Fig.~\ref{manufacturing}(e)). Finally, we used the impulse heat sealer to create the exterior seals (Fig.~\ref{manufacturing}(f)), and we pressurized the pocket using the air tube and airtight cap. 

We placed the incisions approximately 3~cm from the exterior seals---close enough to minimally obstruct the body of the sensor but far enough to fit the components without interfering with the sealing process. After adding extra space in the end pockets for the sensor and the tubing, we spaced the partial seals equally along the length of the pocket to make it a uniform height; more seals with closer spacing gives the inflated pocket a lower profile and higher flexibility, but it slows internal airflow within the sensor, so we used three partial seals. 

\begin{figure}[t]
\centerline{\includegraphics[width = \columnwidth]{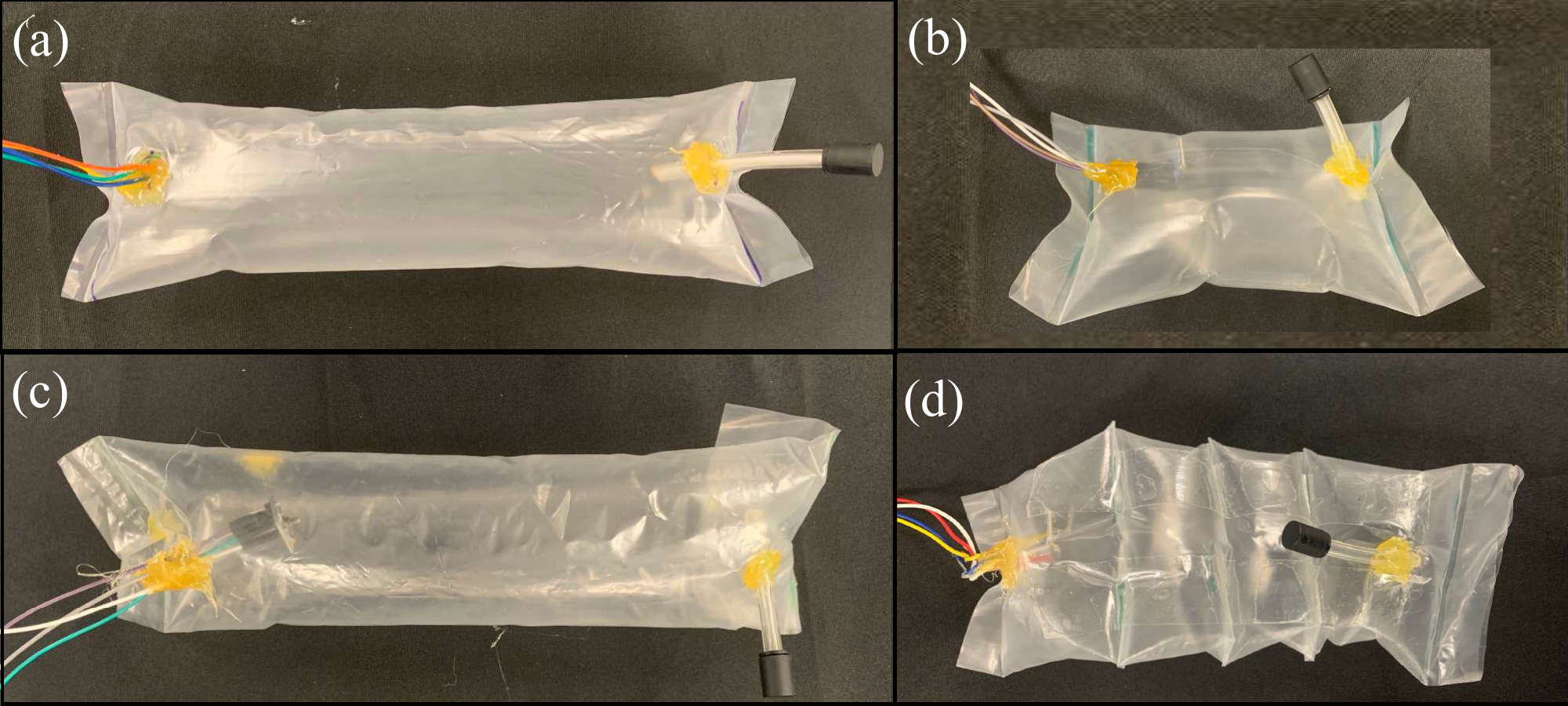}}
\caption{The four sensor geometries used for testing. (a) The control pocket has thickness 0.10~mm and pre-inflated length 27.5~cm. (b) The small pocket has thickness 0.10~mm and pre-inflated length 15~cm. (c) The thin pocket has thickness 0.05~mm and pre-inflated length 27.5~cm. (d) The sealed pocket has thickness 0.10~mm and pre-inflated length 27.5~cm, with sub-pocket pre-inflated lengths of 8~cm, 5.75~cm, 5.75~cm, and 8~cm. All pockets have lay flat diameter 10.2~cm.}
\vspace{-0.5cm}
\label{alltest}
\end{figure}

We fabricated four test sensor geometries (Fig.~\ref{alltest}), which we used to study the effect of various parameters on the sensor's change in internal air pressure when force is applied (Sec.~\ref{Experimental Exploration of Sensor Calibration}). Our control pocket (Fig.~\ref{alltest}(a)), used as a baseline to compare the other pockets against, has a thickness of 0.10~mm and a pre-inflated length (measured between the centers of the two external seals) of 27.5~cm. Our small pocket (Fig.~\ref{alltest}(b)) has a thickness equal to that of the control pocket and a shorter pre-inflated length of 15~cm. Our thin pocket (Fig.~\ref{alltest}(c)) has a lower thickness of 0.05~mm and a pre-inflated length equal to that of the control pocket. Our sealed pocket (Fig.~\ref{alltest}(d)), has a thickness and pre-inflated length equal to that of the control pocket, and the partial internal seals create four subpockets with pre-inflated lengths of 8~cm, 5.75~cm, 5.75~cm, and 8~cm.

\subsection{Integration onto a Vine Robot}

To demonstrate the use of our force sensors in robot control (Sec.~\ref{Demonstration}), we integrated an array of six of the sealed pocket sensors onto a vine robot (Figs.~\ref{glamorshot} and~\ref{demo}). We made the vine robot's main body out of an LDPE tube with 20.3~cm lay flat diameter and 0.10~mm thickness. On the left and right sides of the main body tube, we used double-sided tape (MD 9000, Marker-Tape, Mico, Texas) to attach two series pouch motor actuators fabricated as in~\cite{coad2019vine} out of LDPE tubes with 10.2~cm lay flat diameter and 0.10~mm thickness. We also used double-sided tape to attach three force sensors along the length of each actuator.

To control the speed of growth of the vine robot, we attached the main tube of the robot body to a base as in~\cite{coad2019vine}, i.e., a rigid pressure vessel with an air inlet. Inside the base, the pre-everted vine robot wraps around a spool. To evert the vine robot, the base is pressurized through the air inlet to a pressure higher than that needed to begin growth~\cite{hawkes_soft_2017}, and the spool is controlled using a motor with an encoder to unroll the robot body material. The air pressure inside the base and main body tube is controlled using a closed-loop air pressure regulator (QB3, Proportion-Air, McCordsville, IN) connected to a compressed air source. 
To steer the robot left and right as it navigates along a tabletop, the air pressures inside the actuators on the left and right sides of the robot body are also controlled by the same type of closed-loop pressure regulators.

All of the sensors and actuators are connected to a single microcontroller (Uno, Arduino, Turin, Italy) on which the robot's control code runs. The air pressure sensors inside our force sensors communicate via I2C, and an I2C multiplexer (TCA9548A, Adafruit, New York, NY) allows the Arduino to read all six sensors in sequence. By using up to nine multiplexers, up to 64 air pressure sensors can be read at once.



\section{Experiments and Results} \label{Experimental Exploration of Sensor Calibration}

In this section, we discuss experiments and results studying the relationship between the force applied to our sensors and their measured internal pressure when the parameters of either the contact or the sensors themselves are varied.


\subsection{Experimental Procedure} \label{Testing}

\begin{figure}[t]
\centerline{\includegraphics[width = 0.8\columnwidth]{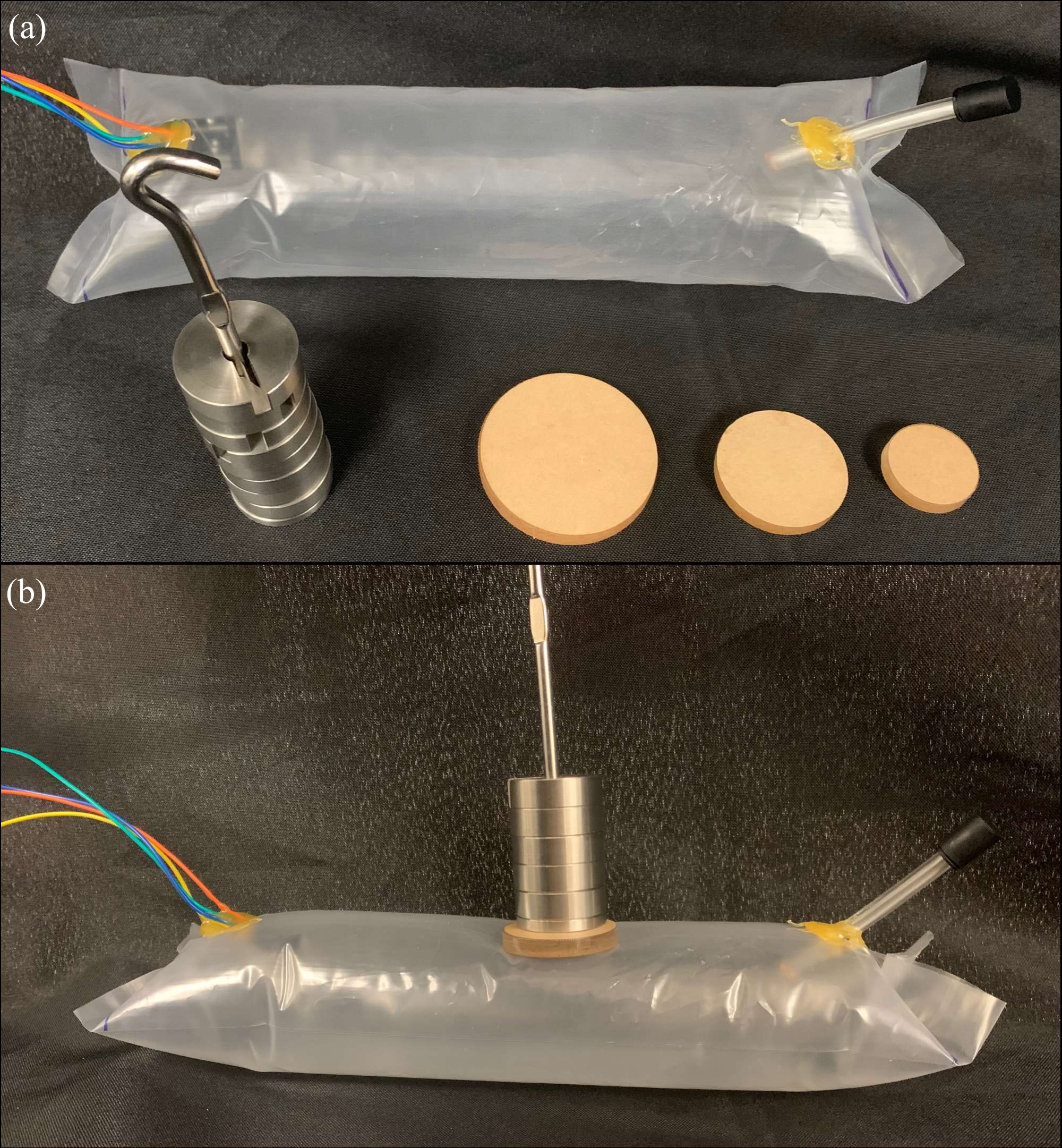}}
\caption{Setup for sensor experimental testing and calibration. 
(a) Components include the sensor itself, a variable weight set,
and three disks of varying areas. (b) We placed the desired weight on one of the disks. We then placed the disk on the sensor in the desired location and recorded the change in internal pressure from its initial value.}
\vspace{-0.5cm}
\label{experimentalsetup}
\end{figure}

Fig.~\ref{experimentalsetup} shows our experimental setup. We placed the sensor to be tested on a table and measured the initial internal pressure using the embedded air pressure sensor. We then placed a low-mass circular acrylic disk on top of the sensor to provide a desired contact area. On top of the disk, we placed the desired weights, and after approximately 5 seconds, we noted the resulting internal pressure. We subtracted off the initial internal pressure to determine the change in pressure for each applied force.

For each experiment, we used weights with masses of 150~g, 300~g, and 450~g, and we conducted three trials for each scenario. We also periodically measured the change in internal pressure with no weight on the pocket, and it was always approximately 0~kPa during our experiments. We chose these particular weights because they allow us to understand the general trends of the sensitivity within a force range that a vine robot might undergo when navigating an environment and contacting obstacles to wrap around.

For most trials, we used a medium-sized disk with contact area 12.5~cm$^2$ and mass 8.75~g, but to study the effect of varying contact area, we used two additional disks: one smaller disk with contact area 6.9~cm$^2$ and mass 4.36~g, and one larger disk with contact area 25~cm$^2$ and mass 17.6~g. Note that in plotting our data, we added the weight of the disk to the weight of the added mass to determine the total force applied to the sensor.

For most trials, we used the control pocket, but to study the effect of varying the sensor length, varying the membrane thickness, or adding the internal partial seals on the sensitivity, we compared the sensitivities of the control pocket with that of the other three pockets described in Section~\ref{Pocket Manufacturing}.

For most trials with the control pocket and all trials with the thin and small pockets, we applied the weight in the middle of the top surface of the sensor. However, to study the effect of varying the contact location, we applied the weight at varying locations. For the control pocket, we also applied the weight on the top of the sensor at 7~cm, 13~cm, and 19~cm from the end containing the pressure sensor, and we also rotated the sensor 90$^\circ$ circumferentially and applied the weight in the middle of the side of the sensor (along the line between the ends of the two end seals). For the sealed pocket, we usually applied the weight to the second sub-pocket from the end closest to the embedded pressure sensor, but we also compared this location with applying the weight to the third sub-pocket.

For most experiments, we used an initial internal pressure of 0.4~kPa above atmospheric. To study the effect of varying the initial internal pressure in the control pocket, we also tested pressures of 0.7~kPa and 1.0~kPa, which we achieved by inflating the control pocket through the air tube to the desired amount and then sealing it with the airtight cap.


After plotting the data for each experiment, we determined the best fit line relating the applied force to the change in pressure. The slope of the line $s$ represents the sensitivity of the sensor and is given by 

\begin{equation}
    s = \frac{P_{sensed}-P_{initial}}{F_{applied}},
\label{pressureleft}    
\end{equation}
where $P_{sensed}$ is the sensed pressure, $P_{initial}$ is the initial pressure, and $F_{applied}$ is the applied force. 

The process of finding $s$ allows us to calibrate the sensor. This calibration can then be used for that sensor to map a sensed pressure change to an applied force. This mapping is given by 

\begin{equation}
    F_{applied} = \frac{P_{sensed}-P_{initial}}{s}.
\label{forceleft}
\end{equation}



\begin{figure*}[t]
\centerline{\includegraphics[width = \textwidth]{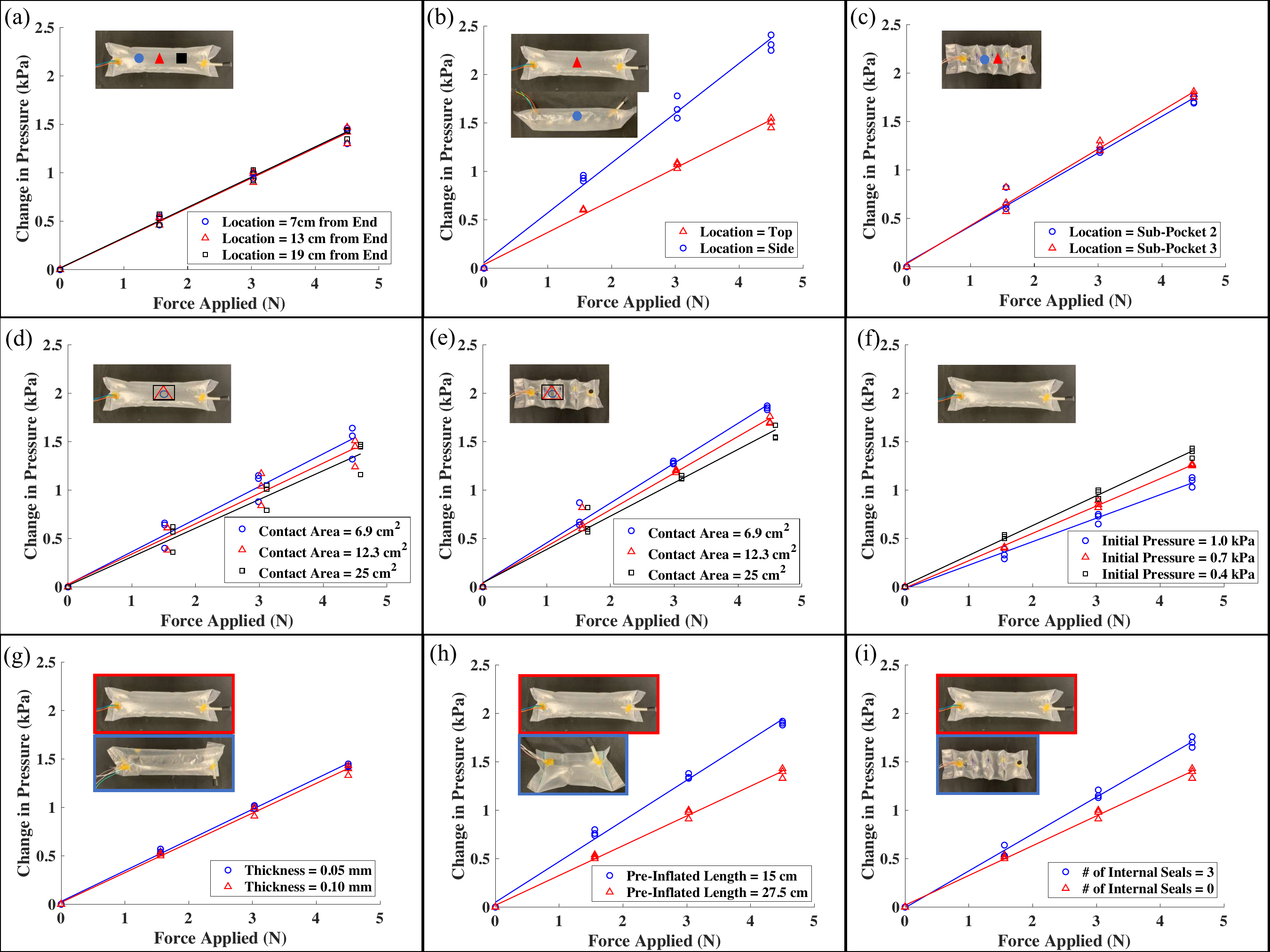}}
\caption{Results for sensor experimental testing. We studied the effect of uncontrollable factors (contact location (a-c) and contact area (d-e)) and controllable factors (initial pressure (f), thickness (g), pre-inflated length (h), and number of interior seals (i)) on the change in internal pressure due to applied force. Raw data is plotted (blue circles, red triangles, or black squares) along with a best fit line for each data set (solid lines).
(a) For the control pocket, there is no detectable relationship between contact location along its length and sensitivity. 
(b) For the control pocket, the sensitivity is significantly greater when contact occurs on the side than on the top.
(c) For the sealed pocket, the sensitivity is very slightly greater when contact occurs at a sub-pocket farther from the end with the pressure sensor than closer to that end.
(d) For the control pocket, the sensitivity increases slightly as contact area decreases. 
(e) For the sealed pocket, the sensitivity also increases as contact area decreases, with slightly greater differences in response between contact areas. 
(f) For the control pocket, the sensitivity increases as initial internal pressure approaches atmospheric. 
(g) 
The thin pocket has a very slightly greater sensitivity than the control pocket.
(h) The small pocket has a significantly greater sensitivity than the control pocket.
(i) The sealed pocket has a greater sensitivity than the control pocket. 
}
\vspace{-0.5cm}
\label{controllablefactors}
\end{figure*}

The results of these experiments are shown in Fig.~\ref{controllablefactors}.

\subsection{Results---Uncontrollable Factors} \label{Uncontrollable Factors}
In this subsection, we present results showing the effect of varying uncontrollable factors---that is, factors that are the properties of contact, not the sensor itself---on the sensitivity of the sensor. Our goal is to create a sensor that behaves consistently when the same force is applied, regardless of these factors.


\subsubsection{Contact Location, Lengthwise} \label{Contact Location: Axial}



As shown in Fig.~\ref{controllablefactors}(a), the sensitivities for the three contact locations along the length of the control pocket are the same to two significant digits with a value of $s$~=~0.31~kPa/N. There is no detectable relationship between location along the length and sensed pressure, which is desirable. 

\subsubsection{Contact Location, Radial} \label{Contact Location: Radial}

As shown in Fig.~\ref{controllablefactors}(b), the sensitivity when force is applied to the side of the control pocket is significantly higher than when force is applied to the top. The slope when force is applied to the top is $s$~=~0.33~kPa/N, which is very similar to the lengthwise location tests. Differences can be attributed to slight differences in initial pressure, which will be discussed in Section~\ref{Internal Pressure}. The slope when force is applied to the side is $s$~=~0.51~kPa/N. This significant pressure difference is likely due to how the material folds. When force is applied to the top, the inflated LDPE holds its shape, but when force is applied along the crease on the side, the inflated LDPE crumples, significantly decreasing the volume of the pocket, and resulting in significantly different sensed pressures, especially at higher forces. To counteract this difference in sensitivity, we suggest attaching the bottom of the pocket to the surface of the flexible robot, so that it is less likely that forces will be applied to the side.

\subsubsection{Contact Location, Seals} \label{Contact Location: Seals} 

As shown in Fig.~\ref{controllablefactors}(c), the sensitivity when force is applied to the third subpocket is slightly higher than when force is applied to the second subpocket. For the second sub-pocket, the slope is $s$~=~0.38~kPa/N, while for the third sub-pocket, the slope is $s$~=~0.39~kPa/N. The increased $s$ for both sub-pockets over the previous lengthwise tests is due to the increased sensitivity seen in sealed pockets, which will be described in Section~\ref{Seals}. The minor difference in sensitivities at the two sub-pockets is likely due to slight differences in the sub-pocket shape as a result of the manufacturing process.

\subsubsection{Contact Area} \label{Contact Area} 


As shown in Fig.~\ref{controllablefactors}(d), the sensitivity for smaller contact areas is slightly higher than that for larger contact areas. For the small disk, the slope is $s$~=~0.34~kPa/N. For the medium disk, the slope is $s$~=~0.31~kPa/N. For the large disk, the slope is $s$~=~0.30~kPa/N. This difference seems to be due to the larger pocket deformation effects the smaller contact areas have. Future work should aim to decrease the effect of contact area on the sensitivity of the pressurized pocket.

\subsubsection{Contact Area, Seals} \label{Contact Area: Seals}

As shown in Fig.~\ref{controllablefactors}(e), the sensitivities for the three contact areas on the sealed pocket follow the same trend as those for the control pocket. For the small disk, the slope is $s$~=~0.41~kPa/N. For the medium disk, the slope is $s$~=~0.38~kPa/N. For the large disk, the slope is $s$~=~0.34~kPa/N. The change in slope with changes in contact area is slightly larger for the sealed pocket than the control pocket, likely because the effect of the larger deformation caused by smaller contact areas is more significant on the sealed pocket.


\subsection{Results---Controllable Factors} \label{Controllable Factors}

In this subsection, we present results showing the effect of varying controllable factors---that is, factors that are properties of the sensor, not the contact---on the sensitivity of the sensor. Our goal is to choose these properties strategically so as to increase the sensitivity of the sensor.

\subsubsection{Initial Pressure} \label{Internal Pressure}
As shown in Fig.~\ref{controllablefactors}(f), a lower initial pressure results in a higher sensitivity than a higher initial pressure. At 0.4~kPa, the slope is $s$~=~0.31~kPa/N. At 0.7~kPa, the slope is $s$~=~0.28~kPa/N. At 1.0~kPa, the slope is $s$~=~0.24~kPa/N. We do not recommend using an initial pressure of 0~kPa, as the pressure response tends to be unreliable, but a pressure around 0.4~kPa has a good combination of sensitivity and reliability.

\subsubsection{Thickness} \label{Thickness}
As shown in Fig.~\ref{controllablefactors}(g), the thin pocket is slightly more sensitive than the control pocket. The slope for the control pocket is $s$~=~0.31~kPa/N, while the slope for the thin pocket is $s$~=~0.32~kPa/N. The thin pocket is more prone to tearing and melting under the hot glue, so we recommend using a thicker pocket when using LDPE plastic.  

\subsubsection{Size} \label{Size}
As shown in Fig.~\ref{controllablefactors}(h), the small pocket is significantly more sensitive than the control pocket. The slope for the small pocket is $s$~=~0.42~kPa/N, while the slope for the control pocket is $s$~=~0.31~kPa/N. This is likely because the change in volume caused by the application of the force is a larger percentage of a small pocket's volume than for a larger pocket, yielding a larger change in pressure. While a larger pocket has more space with which to find forces, a smaller pocket has a greater sensitivity to differentiate between forces. The size best suited for a given robot is dependent on the task.

\subsubsection{Seals} \label{Seals}
As shown in Fig.~\ref{controllablefactors}(i), the sealed pocket is more sensitive than the control pocket. The slope for the sealed pocket is $s$~=~0.38~kPa/N, while the slope for the control pocket is $s$~=~0.31~kPa/N. Thus, in addition to being more flexible and lower profile for integration onto a vine robot, the pocket with seals is also preferable in its sensitivity.


\section{Demonstration} \label{Demonstration}

We conducted a demonstration to showcase the use of our force sensors to enable new capabilities for a vine robot. Our goals were to show that the vine robot is able to grow and steer with the sensors and that the sensors can allow the robot to respond to force in the environment to inform its control. This demonstration is inspired by the motion of natural vines, which search for and wrap around objects in their environment to support themselves and extend their reach. With the addition of force sensing, a vine robot should be able to do the same, provided that the object is large enough that the robot has a tight enough turning radius to continuously wrap around it. 

In our demonstration, the robot is instructed to sweep out a path while searching for contact on its left and right sides. If contact occurs, the robot grows and steers towards the sensed object, attempting to wrap around it. If contact is sustained as the robot grows and subsequent sensors continue to detect contact, the robot continues to wrap around the object. Otherwise, the object is deemed too small, and the robot continues searching for contact elsewhere. 






\begin{figure}[t]
\centerline{\includegraphics[width = \columnwidth]{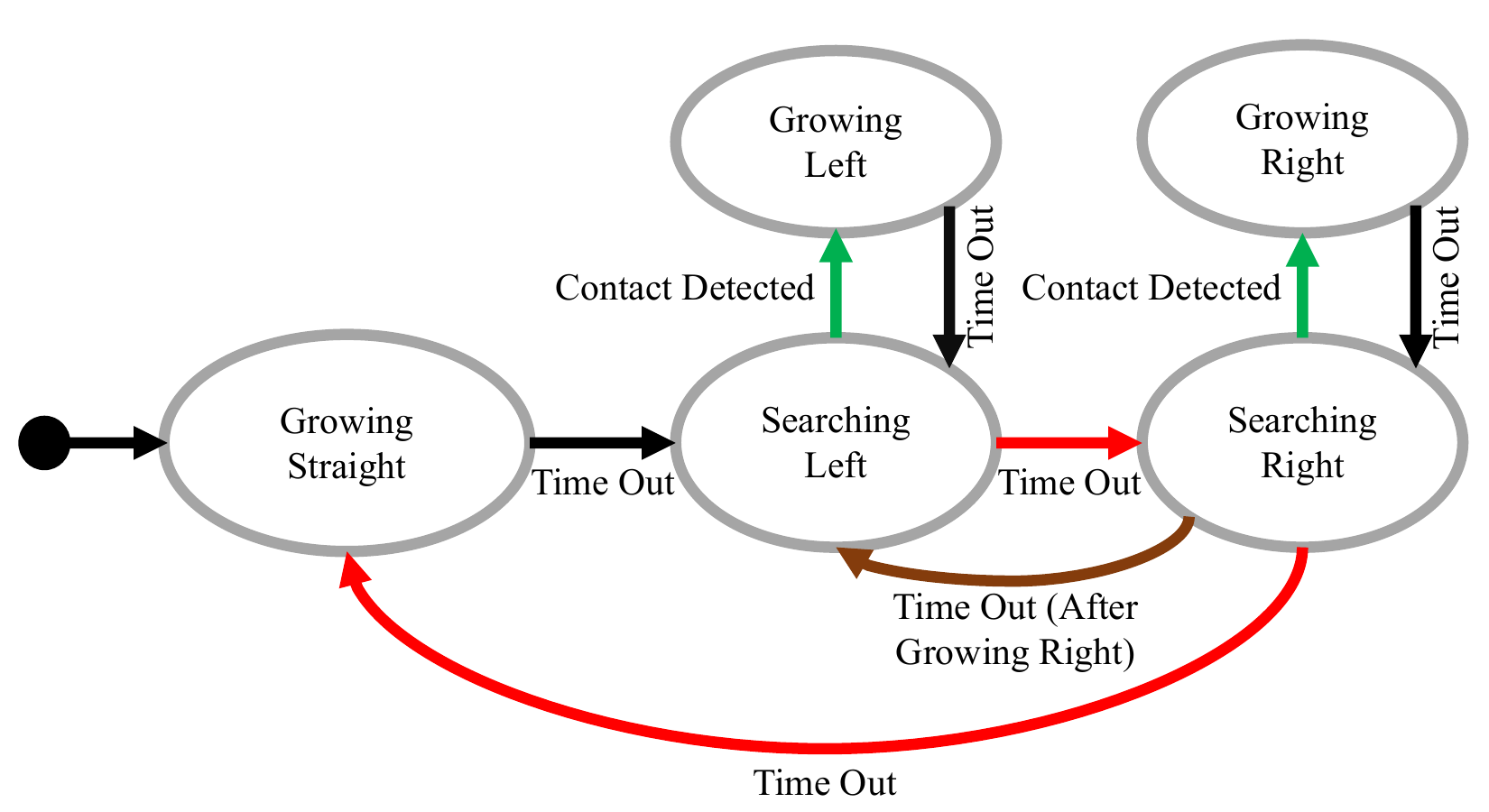}}
\caption{State diagram for contact search algorithm used in the demonstration. Until contact is detected, the robot cycles through the main three states shown across the middle: it grows straight (Growing Straight), then it stops and steers left (Searching Left) and then right (Searching Right) to search for contact. If contact is detected during one of the search states, the robot grows while steering either left (Growing Left) or right (Growing Right) towards the detected contact to attempt to wrap around the object that caused the contact. If continued contact is not detected, the robot goes back to the main three states. Green arrows represent contact detection events, black arrows represent time out events after the robot grows by the desired length, red arrows represent time out events after contact is not detected, and the brown arrow represents a time out event after contact is not detected when there was contact detected to the right in the previous searching stage.}
\vspace{-0.5cm}
\label{alg}
\end{figure}


The robot control uses a state machine (Fig.~\ref{alg}) to achieve this task. There are five states: Growing Straight, Searching Left, Searching Right, Growing Left, and Growing Right. In Growing Straight, the vine robot grows with no steering until a time out of 12~s is reached, which corresponds to growing the length of one pocket. In Searching Left and Searching Right, the robot steers to the left or the right with no growth. During the searching states, the robot checks for contact on the front sensor on the side that it is steering towards. If the threshold pressure value of 1.01~kPa is reached before a time out of 15~s, contact is detected and robot begins to grow towards the sensed contact. Otherwise, the robot searches for contact elsewhere. In Growing Left and Growing Right, the robot steers to the left or the right while it grows the length of one pocket (12~s). Then, it returns to the appropriate sensing state and checks for contact with the newly grown front sensors.

\begin{figure}[t]
\centerline{\includegraphics[width = \columnwidth]{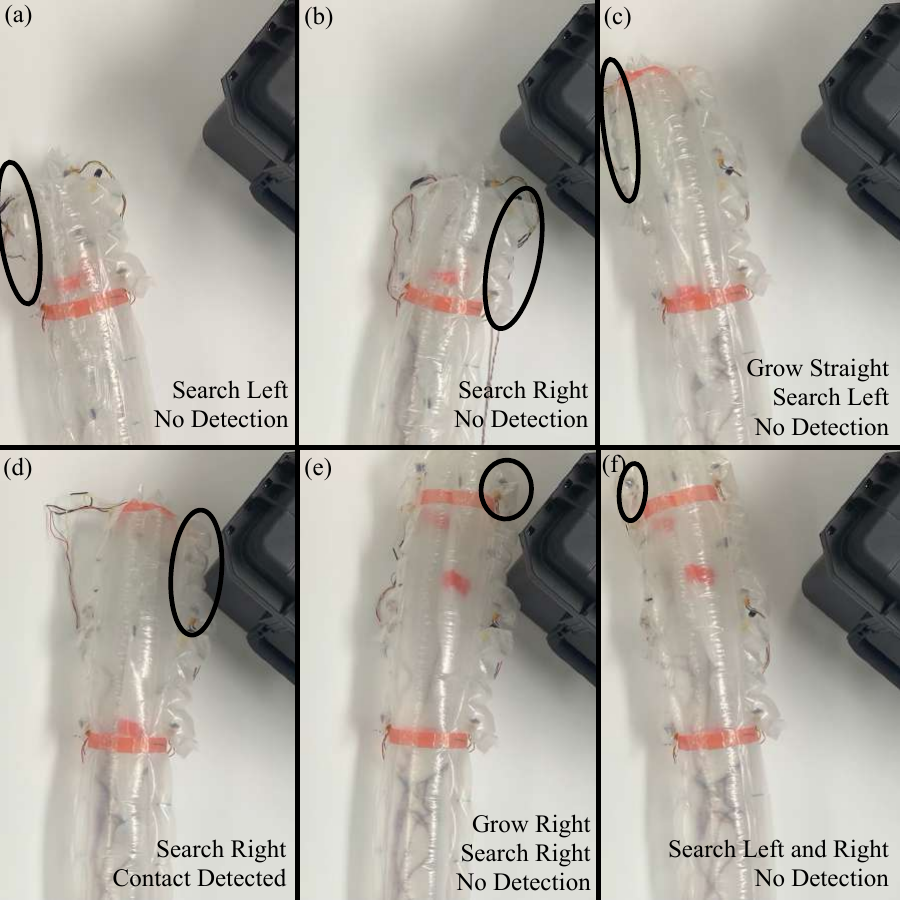}}
\caption{Demonstration of our sensors used to control a vine robot to grow and steer to wrap around an object. (a) The robot begins by steering left, searching for contact, which is not detected. (b) The robot steers right and searches for contact, which is not detected.
(c) The robot grows straight and then repeats the process of searching for contact to the left, which is not detected. (d) The robot searches for contact to the right and detects an object to wrap around. (e) 
The robot continues to steer to the right while it grows, attempting to wrap around the object. The newly everted sensor at the tip searches for contact to the right to determine whether contact continues, which is not detected. (f) The object is too small for the robot to continuously wrap around, so the robot leaves the object and continues searching.}
\vspace{-0.5cm}
\label{demo}
\end{figure}

Fig.~\ref{demo} shows the successful demonstration. The vine robot was able to steer left and right with the sensors and grow to expose more sensors. It stopped steering once a sensor sensed contact and was able to respond to that contact by growing while maintaining the current steering pressure. Once the vine robot grew the length of another pocket, the new front sensor did not sense any contact, so the vine robot stopped steering and continued its search for a large enough object.

\section{Conclusion and Future Work} \label{Conclusion and Future Work}

We presented a novel force sensor made of an airtight pressurized pocket containing an embedded air pressure sensor. This force sensor is soft and can attach to the exterior of a flexible robot such as a vine robot while still allowing the vine robot to grow and steer. We demonstrated that this sensor can be used to adjust the control of a vine robot in response to contact.
We measured the effect of both uncontrollable (contact location, contact area) and controllable (initial pressure, thickness, size, and number of seals) factors on the sensitivity of the sensor. 
We found that the location of contact along the robot's body length has a fairly low impact on a pocket, while radial location has the greatest impact. Contact area also has a noticeable but smaller impact. For the controllable factors, a lower initial pocket pressure, a thinner pocket, a smaller pocket, and a pocket with seals all provide higher sensitivities to varying degrees, with pocket size having the biggest impact. 

Future work will quantify the accuracy of our force sensor in detecting applied forces under various conditions, such as loading under different directions while attached to a vine robot. We will quantify and explore how to reduce both the effect of the addition of these sensors on the ability of a vine robot to grow and steer and the effect of the growth and steering of a vine robot on the function of the sensors. We will also investigate whether hysteresis and relaxation are present in the sensor's sensitivity and explore how to reduce them, and we will explore adjustments to the sensor design to reduce the impact of the uncontrollable factors on the sensitivity. Finally, we will explore the use of these sensors in more scenarios, towards our goal of improving the ability of robots to interact intelligently with the physical world.

\bibliographystyle{IEEEtran}
\bibliography{References}

\end{document}